\documentclass{article}

    \usepackage[preprint]{neurips_2024}

\usepackage[utf8]{inputenc} 
\usepackage[T1]{fontenc}    
\usepackage{hyperref}       
\usepackage{url}            
\usepackage{booktabs}       
\usepackage{amsfonts}       
\usepackage{nicefrac}       
\usepackage{microtype}      
\usepackage{xcolor}         
\usepackage{subfigure}
\usepackage{graphicx}
\usepackage{amsthm}
\usepackage{enumitem,kantlipsum}
\newtheorem{definition}{Definition}

\title{There is More to Graphs than Meets the Eye: Learning Universal Features with Self-supervision}

\author{%
  Laya Das$^*$\\
  ETH Zurich \\
  8092 Zurich, Switzerland\\
  \texttt{laydas@ethz.ch}\\
  \And
  Sai Munikoti$^*$
  \\
  Pacific Northwest National Lab \\
  Richland \\
  WA, USA\\
  \texttt{sai.munikoti@pnnl.gov}
  \And
  Nrushad Joshi$^\dagger$ \\
  Indiana University Bloomington \\
  Bloomington \\
  IN, USA\\
  \texttt{nrujoshi@iu.edu}
  \And
  Mahantesh Halappanavar \\
  Pacific Northwest National Lab \\
  Richland \\
  WA, USA\\
  \texttt{mahantesh.halappanvar@pnnl.gov} \\
}

\begin{document}

\maketitle
\def\thefootnote{*}\footnotetext{These authors have contributed equally to this work}
\def\thefootnote{$\dagger$}\footnotetext{Work done at PNNL}
\def\thefootnote{\arabic{footnote}}

\begin{abstract}
We study the problem of learning features through self-supervision that are generalisable to multiple graphs. State-of-the-art graph self-supervision restricts training to only one graph, resulting in graph-specific models that are incompatible with different but related graphs. We hypothesize that training with more than one graph that belong to the same family can improve the quality of the learnt representations. However, learning universal features from disparate node/edge features in different graphs is non-trivial. To address this challenge, we first homogenise the disparate features with graph-specific encoders that transform the features into a common space. A universal representation learning module then learns generalisable features on this common space. We show that compared to traditional self-supervision with one graph, our approach results in (1) better performance on downstream node classification, (2) learning features that can be re-used for unseen graphs of the same family, (3) more efficient training and (4) compact yet generalisable models. We also show ability of the proposed framework to deliver these benefits for relatively larger graphs. In this paper, we present a principled way to design foundation graph models that learn from more than one graph in an end-to-end manner, while bridging the gap between self-supervised and supervised performance.
\end{abstract}

\section{Introduction}
Self-supervised learning (SSL) aims to learn generalisable representations from large corpora of unlabelled datasets that can be used for several downstream tasks \citet{kolesnikov2019revisiting,he2022masked}. Recent progress in graph SSL has pushed the state-of-the-art (SOTA) performance on several benchmark datasets and tasks \citet{xiao2022decoupled,jin2022automated,liu2022graph,balestriero2023cookbook}, at times outperforming suprvised baselines \citet{}. These methods typically focus pre-training to only one dataset, with one \citet{liu2022graph} or many \citet{jin2022automated} pre-training tasks, effectively learning graph-specific representations and models that are incompatible with other related graphs. This is equivalent to training a masked autoencoder model with ImageNet, which is incompatible with MS-COCO dataset. Thus, unlike SSL in natural language processing and computer vision, current graph SSL suffers from the lack of a framework that can simlutaneously learn from multiple related datasets, and in the true spirit of SSL, exploit large corpora and produce models that can work with different graphs.

SOTA graph SSL frameworks that train with only one graph exhibit crucial deficiencies. First, each model learns a distinct set of parameters, independent of other similar datasets, precluding the use of shared parameters that could lead to learning universal features. This hampers generalizability of the resulting models, and as shown in this work, can also lead to poor performance on downstream node classification. Second, owing to the disparate node and edge features of different datasets, SOTA graph SSL models are not compatible with other datasets. So, availability of new datasets mandates building new models from scratch, and one cannot leverage previously learnt representations to inform the training process and reduce the computational load. In other words, SOTA graph SSL models do not exhibit adaptability. Finally, training a separate model for each dataset increases the computational cost of self-supervision and requires proportionally more storage, adding to the cost of SSL. While the current training costs for graph neural networks are much smaller compared to language and vision models, the increasing trend of graph dataset sizes can elevate this cost in the future. Thus, it is important to develop a combined learning framework to address this gap and enable learning simultaneously from multiple graphs, paving the way for more capable SSL and {\em foundation graph models}.

Learning universal representations across graphs poses an important challenge of disparate node and edge features for different graphs. Node features of different graphs typically exhibit different dimensionalities that prevents them from being processed together. For example, the features of \texttt{Cora} and \texttt{Citeseer} have different dimensionalities even when both are citation networks. In datasets where the dimensions match, the individual features of different graphs can be obtained through different processes (e.g., average embedding of words in abstract, or in the entire article), bearing different meanings, that hinder unified processing of these features. Thus, it is imperative for a universal SSL approach to be able to accommodate this diversity, and treat disparate node and edge features in a unified manner. Along similar lines, there has been an increased interest in developing models that can handle data of different modalities, and learn features from different sources of data, such as videos and text, through modular structures and carefully crafted embeddings \citet{gao2020multi,akbari2021vatt}. These foundation multi-modal approaches transform multi-modal data into a common representation space to learn better and robust features. Such an approach has met with incredible success \citet{lu2022multimodal, lu2022unified, wang2022ofa, xu2023mplug}, and is paving the way towards artificial general intelligence \citet{fei2022towards}. Inspired by the success of these models, our work aims to develop a first-of-its-kind universal learning approach for graphs and investigate if the resulting models exhibit better performance in downstream tasks.

\begin{figure}
    \centering
    \subfigure[Model architecture for universal self-supervision]{
        \includegraphics[width=\textwidth]{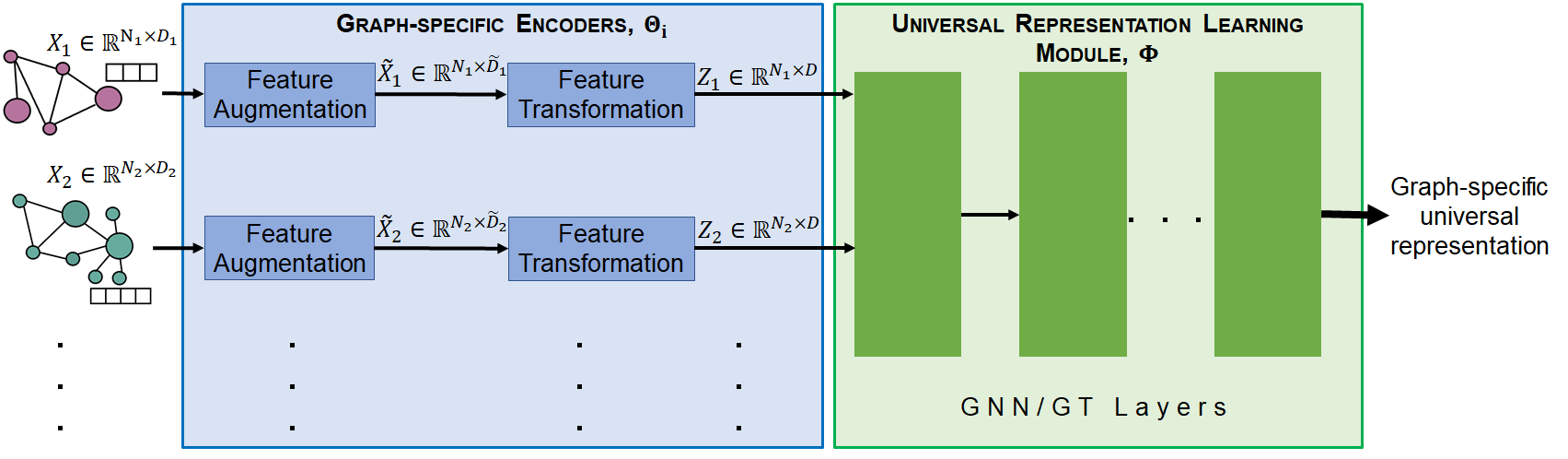}
        \label{fig:model}
    }
    \subfigure[Pre-training]{
        \includegraphics[width=0.45\textwidth]{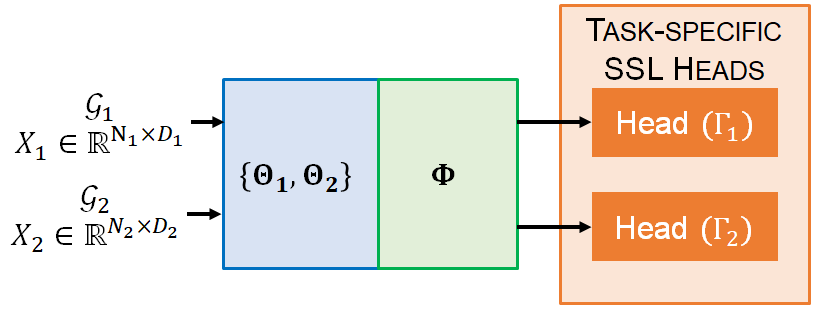}
        \label{fig:pretraining}
    }
    \subfigure[Fine-tuning]{
        \includegraphics[width=0.45\textwidth]{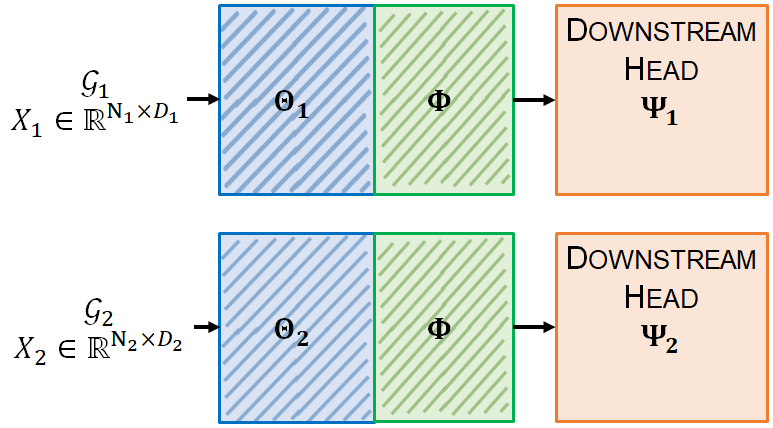}
        \label{fig:finetuning}
    }
    \caption{Universal Self-supervised Learning (U-SSL) across graphs. (a) Model architecture for U-SSL with graph-specific ($\mathbf{\Theta_i}$) and universal ($\mathbf{\Phi}$) parameters. (b) U-SSL pre-training with two graphs, $\mathcal{G}_1$ and $\mathcal{G}_2$. (c) Downstream task learning for individual graphs. Hatched boxes represent frozen parameters ($\mathbf{\Theta_i,\Phi})$, and shaded boxes represent learnable parameters ($\mathbf{\Psi_i}$).}
    \label{fig:framework}
\end{figure}

\textbf{Contributions:}
Our approach is rooted in the observation that graphs belonging to the same family are known to exhibit universal patterns \citet{sharan2005conserved,wang2021science}. The proposed framework, called Universal Self Supervised Learning (U-SSL) leverages these universal patterns from graphs in a family and explicitly addresses the challenges with SSL discussed above. In this article, we

\begin{enumerate}
    \item present a universal representation learning framework through self-supervision for graphs (U-SSL). The framework is modular and allows training with arbitrary choices for the (1) number of training graphs and (2) number and type of pretext tasks.
    \item construct U-SSL models with (1) graph-specific encoders that accommodate the disparity of node features of different graphs, and (2) a universal module that learns representations generalisable to all graphs used during training. The model allows end-to-end training, thus simultaneously learning both graph-specific and universal parameters. The model is constructed in a modular way so that it can be made to work with new graphs -- as and when they are available -- by simply adding a new graph-specific module and re-training only this module.
    \item demonstrate the superiority of U-SSL models with a use-case study on citation networks with (1) better efficacy (\textbf{$\mathbf{1}$ to $\mathbf{8}$ points improvement} in downstream node classification accuracy), (2) better efficiency ($\mathbf{6\%}$ \textbf{reduction in training time} per epoch for five datasets) and (3) lower model sizes ($\mathbf{60\%}$ \textbf{savings in parameter count}) compared to SSL models.
    \item demonstrate \textbf{adaptability of U-SSL models to an unseen dataset}, a feature not provided by any SOTA graph SSL framework.
    \item show \textbf{generalisability} of the framework with multiple pretext tasks and graph families.
\end{enumerate}

\section{Related Work}

\paragraph{Graph neural networks and graph transformers}
Graph neural networks have been extremely successful in learning representations from graph-structured data, and solving challenging problems in applications including neuroscience \citet{wein2021graph}, medicine \citet{bongini2021molecular}, optimization \citet{schuetz2022combinatorial} and many more.  Most GNN architectures can be broadly categorized as message passing networks, that operate in two stages, i.e., aggregation and combination, with different architectures performing these steps in different ways. One of the earliest GNNs generalized the convolution operation to graph-structured data, and proposed the Graph Convolutional Network (GCN) \citet{kipf2016semi}. This was followed by an explosion of GNN models, such as GraphSAGE \citet{hamilton2017inductive}, Graph Attention Networks (GAT) \citet{veličković2018graph} and Graph Isomorphism Networks (GIN) \citet{xu2018how} that crafted different aggregation and combination operations to capture different relationships in graphs. For instance, GAT uses an attention mechanism for aggregation to assign different weights to different nodes in a neighborhood, allowing the model to focus on the most relevant nodes for a given task, and obtain better performance than GCN that uses convolution for aggregation.

Message passing networks (MPNs) suffer from fundamental limitations, e.g., over-smoothing \citet{Oono2020Graph}, over-squashing \citet{alon2021on} and expressive limits \citet{morris2019weisfeiler}, that are addressed with graph transformers \citet{rampavsek2022recipe}. GTs make use of positional or structural embeddings along with global attention mechanisms to learn both local and global features and thus address the limitations of MPNs \citet{rampavsek2022recipe}. Several GT architectures have been proposed for homogeneous graphs \citet{yun2019graph,kreuzer2021rethinking}, heterogeneous graphs \citet{hu2020heterogeneous} and hyper-graphs \citet{kim2021transformers}. GTs, however, relatively require more training data and do not generalize well to unseen graphs \citet{zhao2021gophormer,chen2023tokenized}.

\paragraph{Graph representation learning with self-supervision} SSL learns generic representations as opposed to task-specific representations in supervised learning. There are several SSL methods on graphs including Deep Graph Infomax \citet{velickovic2019deep} and Auto-SSL \citet{jin2022automated}, as well as reviews \citet{jin2022automated,xie2022self,liu2022graph} that summarize the state-of-the-art. Graph SSL has been performed with contrastive as well as predictive learning tasks \citet{xie2022self}. While the former aim to learn representations by distinguishing positive and negative samples, the latter seek to predict the values of masked or corrupted nodes or edges. For instance, \citet{velickovic2019deep} adopt contrastive learning and maximize mutual information between local patches of a graph and the global graph representation to learn node representation. \citet{rong2020self} apply SSL to molecular graphs to learns representations by predicting masked nodes and edges. There are several SSL tasks such as node attribute masking, graph structure prediction, and graph context prediction, which can be used to learn representations in a self-supervised manner.

The majority of graph self-supervision is performed with one graph and one SSL task. \citet{jin2022automated} proposed a mechanism to automate self-supervision with multiple tasks, by adaptively weighing the losses of different tasks during training. Their framework, named Auto-SSL, extended SSL to include multiple tasks during training. However, all SOTA graph SSL methods use only one graph/dataset to learn representations prior to downstream task learning. We address this gap, and a framework to learn universal representations across different graphs -- of a certain family.

\section{Learning Universal Features with Graph self-supervision}
In this section, we describe the problem formulation and our hypothesis on improving graph representation learning, followed by the construction, pre-training and fine-tuning of U-SSL models.

\subsection{Problem Formulation and Hypothesis}
We consider $N$ graphs $\{\mathcal{G}_i\}_{i=1}^N$, with each graph represented as a tuple of nodes $\mathcal{V}_i$ and edges $\mathcal{E}_i$, $\mathcal{G}_i=\left(\mathcal{V}_i, \mathcal{E}_i\right)$ such that $|\mathcal{V}_i|=N_i$ and $\mathcal{E}_i\subseteq\mathcal{V}_i\times\mathcal{V}_i$. Let $\mathbf{A_i}\in\{0,1\}^{N_i\times N_i}$ and $\mathbf{X_i}\in\mathbb{R}^{N_i\times D_i}$ represent the adjacency matrix and node feature matrix of $\mathcal{G}_i$, respectively. Let $\mathcal{L}_{SSL, i}$ denote the pretext task loss for graph $\mathcal{G}_i$. We then provide the definition of SSL, as studied in the current literature as:
\begin{definition}
    For graph $\mathcal{G}_i$, the problem of self supervised learning is to learn an encoder $f_i\left(\mathbf{X_i},\mathbf{A_i};\mathbf{\Theta_i}\right)$ by minimizing the loss $\mathcal{L}_{SSL, i}$ such that the learnt representations can be used to solve downstream learning tasks for $\mathcal{G}_i$.
\end{definition}
We extend this definition to the problem of learning universal features with self-supervision (U-SSL) as follows:
\begin{definition}
    For graphs $\{\mathcal{G}_i\}$, the problem of universal self-supervision is to learn an encoder $f\left(\{\mathbf{X_i}\},\{\mathbf{A_i}\};\{\mathbf{\Theta_i}\},\mathbf{\Phi}\right)$ by minimizing the loss $\sum_{i=1}^N\mathcal{L}_{SSL, i}$ such that the learnt features can be used to solve downstream tasks for $\{\mathcal{G}_i\}$.
\end{definition}
The U-SSL model can take as input, disparate features from different graphs, and learn universal features that are common to all the datasets, thereby generalizing well to these datasets, and potentially also to other similar datasets. We note that different graphs have different node feature sizes, i.e., in general, $D_i\neq D_j$ for $i\neq j$. This necessitates that there be parts of the encoder $f$ dedicated to different graphs, with graph-specific parameters $\mathbf{\Theta_i}$, in addition to the universal parameters $\mathbf{\Phi}$.

Let us denote the representations learnt for graph $\mathcal{G}_i$ with SSL as $\mathbf{H_i^s}$, and those learnt with U-SSL as $\mathbf{H_i^u}$, i.e.,
\begin{eqnarray}
    \mathbf{H_i^s}&=&f_i\left(\mathbf{X_i},\mathbf{A_i};\mathbf{\Theta_i}\right),\\
    \mathbf{H_i^u}&=&f\left(\mathbf{X_i},\mathbf{A_i};\mathbf{\Theta_i},\mathbf{\Phi}\right).
\end{eqnarray}
Our hypothesis is that U-SSL can learn representations that are better than those learnt with SSL, in terms of solving a downstream task, e.g., node classification, for graphs $\{\mathcal{G}_i\}_{i=1}^N$. Let us denote the downstream task head for graph $\mathcal{G}_i$ as $h_i(\cdot;\mathbf{\Psi_i})$, and let $\mathcal{M}$ be a metric such that higher values of $\mathcal{M}$ represent better performing models. Then, our hypothesis can be formally stated as:
\begin{eqnarray}
    \mathcal{H}:& \mathcal{M}\left(h_i\left(\mathbf{H_i^u};\mathbf{\Psi_i^u}\right)\right)>\mathcal{M}\left(h_i\left(\mathbf{H_i^s};\mathbf{\Psi_i^s}\right)\right).
\end{eqnarray}
Here, the superscripts in $\mathbf{\Psi_i}$ signify that the parameters learnt during fine-tuning of SSL and U-SSL models will be different for the same downstream task head $h_i$.

In formulating our hypothesis, we view a graph $\mathcal{G}_i$ as being an instance of some underlying real-life phenomenon. For instance, \texttt{Cora}, and \texttt{Citeseer}, are two instances of the same underlying real-life phenomenon, i.e., citation among research articles. Learning representations with SSL allows one to extract patterns from only one instance of the underlying phenomenon, while U-SSL allows learning from multiple instances, and hence, observing the underlying phenomenon through multiple lenses. As a result, U-SSL allows learning representations that are fundamental to the underlying mechanism, and is not restricted to the patterns observed in one instance. This can lead to learning more generic features, and hence better downstream performance with U-SSL.

\subsection{Graph-specific Encoder}
The core idea of U-SSL is to learn representations that are generalizable across multiple graphs. This entails processing node features from different graphs in a unified pipeline. However, node (and edge) features of different graphs are obtained with different algorithms, and are typically disparate, i.e., (a) they do not have the same dimensionality, and (b) the entries of feature vectors can bear different meanings for different graphs, even if they have the same dimensionality. It is thus imperative to first homogenize the node (and edge) features of different graphs from their original disparate spaces (of dimension $D_i$) to a common space (of dimension $D$) for processing by the rest of the model. We therefore need graph-specific encoders, represented as $g_i(\cdot;\mathbf{\Theta_i})$ for graph $\mathcal{G}_i$. The encoder $g_i$ can be any neural network module, e.g., GCN layers, linear layers, etc. that transforms the feature vectors into $\mathbb{R}^D$, and can additionally involve pre-processing steps such as node feature augmentation to enrich the feature vectors. In our proposed framework, we include feature augmentation ($F_A$) followed by feature transformation ($F_T$), that transform the node features $\mathbf{X_i}\in\mathbb{R}^{N_i\times D_i}$ to $\tilde{\mathbf{X}}_\mathbf{i}\in\mathbb{R}^{N_i\times \tilde{D}_i}$ to $\mathbf{Z_i}\in\mathbb{R}^{N_i\times D}$:
\begin{eqnarray}
    \tilde{\mathbf{X}}_\mathbf{i}&=&F_A\left(\mathbf{X_i}\right),\\
    g_i(\mathbf{X_i};\mathbf{\Theta_i})&=&\mathbf{Z_i}=F_T\left(\tilde{\mathbf{X}}_\mathbf{i};\mathbf{\Theta_i}\right),\\
    &=&F_T\left(F_A\left(\mathbf{X_i}\right);\mathbf{\Theta_i}\right).
\end{eqnarray}
In general, the functions $g_i$, $F_A$ and $F_T$ also take the adjacency matrix $\mathbf{A_i}$ as input, which is omitted here for brevity. The output of the graph-specific encoders $\mathbf{Z_i}$ represents the graph-specific homogenized features that exist in $\mathbb{R}^D$, $\forall \mathcal{G}_i$ and whose individual entries represent the same quantity across all graphs. In an $N$-graph application, the U-SSL model will be constructed with $N$ different graph-specific encoders, as shown in Fig. \ref{fig:framework}.

\subsection{Universal Representation Learning Module}
The universal representation learning (URL) module aims to learn features that are generic to all $N$ graphs used during pre-training, and thus capture patterns that are fundamental to the underlying process. It takes in the homogenized node features $\mathbf{Z_i}$ from all graphs, and learns the graph-specific universal features, denoted as $\mathbf{H_i^u}$ for graph $\mathcal{G}_i$. The URL module, denoted as $g(\cdot,\mathbf{\Phi})$ for all graphs $\{\mathcal{G}_i\}$ can be any neural network module, e.g., GNN layers or GT blocks, and can be expressed as:
\begin{equation}
    \mathbf{H_i^u}=g(\mathbf{Z_i};\mathbf{\Phi}), \forall i \in [1,N].
\end{equation}

These features are graph-specific since they are obtained from the homogenized node features of a particular graph, and at the same time universal, because they are learnt by minimizing the collective loss accrued for all graphs. A U-SSL model for $N$ graphs is thus constructed with $N$ graph-specific encoders and one universal representation module, as shown in Fig. \ref{fig:model}. This modular nature of the model architecture allows adding as many graph-specific encoders as desired, and simultaneously processing disparate node features, thus facilitating end-to-end training of the model. In addition, this modular nature renders adaptability to the model, wherein a new graph-specific encoder can be introduced to the model without having to alter the rest of the model structure, and re-train, or continue training with the new dataset.

\subsection{Pre-training and fine-tuning U-SSL models}
Pre-training models with SSL involves selecting one or more pre-training task (also referred to in the literature as pretext tasks), typically depending on the type of downstream task, and appending a model with heads to learn the different tasks. Pre-training of U-SSL models is also performed in a similar vein, i.e., by using the U-SSL model with $N$ graph-specific modules, one universal representation module and one or more task-specific heads. Let $\mathbf{\Gamma}$ represent the task-specific head parameters for a pretext task and $\mathcal{L}_{SSL, i}$ represent the loss for $i^{th}$ graph. Then, the total loss for $N$ graphs can be expressed as:
\begin{equation}
    \mathcal{L}_{USSL}=\sum_{i=1}^N\mathcal{L}_{SSL, i}\left(\mathbf{X_i}, \mathbf{A_i}; \mathbf{\Theta_i}, \mathbf{\Phi}, \mathbf{\Gamma}\right). \label{eq:loss}
\end{equation}
The total loss $\mathcal{L}_{USSL}$ is used to simultaneously learn the parameters $\{\mathbf{\Theta_i}\}$, $\mathbf{\Phi}$ and $\mathbf{\Gamma}$ in an end-to end manner. The U-SSL loss can also be generalised to any number of tasks, which is discussed later. At the time of downstream task learning, new heads are appended to the model, parameterized in $\mathbf{\Psi_i}$, which are learnt separately for each graph by keeping the learnt parameters $\{\mathbf{\Theta_i}\}$, and $\mathbf{\Phi}$ unchanged. The pre-training and fine-tuning of U-SSL models are depicted in Fig. \ref{fig:pretraining} and \ref{fig:finetuning}, respectively.

\section{Experiments}
In our main study, we consider $6$ citation network datasets, i.e., \texttt{CoraFull}, \texttt{Cora-ML}, \texttt{DBLP}, \texttt{Citeseer}, \texttt{PubMed} and \texttt{OGBN-arxiv}. We use three GNN architectures (GCN, GraphSAGE and GAT) and two GT architectures (NAGphormer and GTX) for the URL module. An embedding dimension of $256$ is used for all models, with $3$ GNN layers and $4$ GT layers with $8$ attention heads in each layer. For the NAGphormer model, we employ Laplacian position embedding of the nodes (of size $15$) to additionally augment node features with structural information ($F_A$), and obtain the augmented node features of dimension $\tilde{D}_i=D_i+15$ for graph $\mathcal{G}_i$. The graph-specific encoders are linear projections ($F_T$) from the augmented node feature dimension $\tilde{D}_i$ to $256$ for graph $\mathcal{G}_i$. We consider only one pretext so that the U-SSL model has $1$ URL module and $1$ task-specific head. The choice of the self-supervision task in our study is guided by the downstream task. Since we are interested in learning features for node classification, we use the pair-wise attribute similarity (\textbf{PairSim}) self-supervision task in our study. This task learns an encoder to differentiate between similar and dissimilar nodes, posed as a two-class classification problem. We use one fully connected layer to learn this task. We demonstrate the superiority of the features learnt with U-SSL by evaluating and comparing the performance of models obtained with SSL, U-SSL and supervised learning on node classification for all the graphs. We further train $10$ instances of these models for the downstream task to account uncertainty and report the mean and standard deviation of classification accuracy for each experiment. The implementation details are provided in Supplementary material (Appendix \ref{app:implementation}).

In additional experiments, we consider multiple pretext tasks (with citation networks) and families (co-purchase networks and social networks) with the above construction.

\section{Results}

\subsection{There is more to graphs than meets the eye}
We present the advantages of U-SSL over SSL in terms of four aspects: (i) {\em efficacy}, i.e., improvement in performance compared to SSL, which enables bridging the gap between supervised and self-supervised performance, (ii) {\em efficiency}, i.e., reduction in training time compared to SSL, (iii) {\em scalability}, i.e., delivering efficacy and efficiency for larger datasets, and (iv) {\em adaptability}, i.e., the ability to leverage representations learnt through U-SSL on a set of datasets, to learn downstream tasks on new datasets. In this section, we present the performance with NAGphormer, and report the results for all other architectures in Supplemental material (Appendix \ref{app:ablation}).

\paragraph{Efficacy:} The node classification accuracy of supervised baseline, SSL and U-SSL models for \texttt{CoraFull}, \texttt{Cora-ML}, \texttt{DBLP}, \texttt{Citeseer} and \texttt{PubMed} is listed in Table \ref{tab:citation_acc}. The U-SSL models outperform the corresponding SSL models, delivering between $1\%$ and $4\%$ improvement in mean accuracy for these datasets. U-SSL provides a performance gain of $1\%$ for \texttt{CoraFull}, $2\%$ for PubMed, $3\%$ for Citeseer, and $4\%$ for \texttt{CoraML} and \texttt{DBLP}. We note that \texttt{CoraFull} has a large number of classes ($70$) and a large number of nodes ($19,793$), resulting in a more difficult classification task. Nevertheless, the U-SSL model still delivers $1\%$ improvement in accuracy for this dataset. Further, the U-SSL model matches the supervised performance for \texttt{DBLP} and \texttt{PubMed} datasets, clearly demonstrating the advantage of U-SSL over SSL. These results support our hypothesis, and demonstrate that \textit{there is more to graphs than can be learnt with plain SSL, and learning universal representations across graphs with U-SSL can bridge the gap between supervised and self-supervised performance.} In addition, we note that the total number of parameters for the five SSL models ($\mathbf{\{\Theta_i\}}$, $\mathbf{\Phi}$) is $14,390,650$, which is $2.46$ times $5,831,29$ parameters for the U-SSL model trained with the five datasets.
\begin{table}[]
    \centering
    \caption{Node classification accuracy of supervised baseline, SSL and U-SSL models. Entries in boldface represent best performance out of SSL and U-SSL. Underlined entries represent U-SSL models that match supervised baseline performance.}
    \label{tab:citation_acc}
    \begin{tabular}{cccc}
        \toprule
         Dataset & Baseline & SSL & U-SSL \\
         \hline
         CoraFull & $0.70 \pm 0.007$ & $0.59 \pm 0.003$ & $\mathbf{0.60} \pm 0.003$ \\
         Cora-ML & $0.87 \pm 0.004$ & $0.80 \pm 0.002$ & $\mathbf{0.84} \pm 0.001$ \\
         DBLP & $0.83 \pm 0.008$ & $0.79 \pm 0.001$ & $\mathbf{\underline{0.83}} \pm 0.001$ \\
         Citeseer & $0.94 \pm 0.003$ & $0.83 \pm 0.001$ & $\mathbf{0.86} \pm 0.002$ \\
         PubMed & $0.87 \pm 0.008$ & $0.85 \pm 0.002$ & $\mathbf{\underline{0.87}} \pm 0.001$ \\
         \bottomrule
    \end{tabular}
\end{table}

\paragraph{Efficiency:} We observe that the number of epochs for convergence of SSL and U-SSL models at the time of pre-training are comparable for all datasets. We therefore report the efficiency in terms of training time per epoch, which is $0.663$ seconds for the five SSL models combined. The U-SSL model exhibits a training time per epoch of $0.609$ seconds, which is a $6\%$ decrease in the total training time of the model.
Thus, in addition to better performance, \textit{U-SSL provides an efficient framework for self-supervised graph representation learning across multiple datasets.}


\paragraph{Scalability:} To show that the benefits of U-SSL can still be reaped for relatively larger datasets, we include the \texttt{OGBN-arxiv} dataset and train the model with $6$ datasets. The supervised baseline model achieves an accuracy of $0.61 \pm 0.007$, while the SSL model provides an accuracy of $0.46 \pm 0.003$ for the \texttt{OGBN-arxiv} dataset. The U-SSL model achieves an accuracy of $0.54 \pm 0.002$, delivering an improvement of $8\%$ in classification accuracy compared to the SSL model. This is a significant gain in performance for a dataset that is much larger than the graphs reported in Table \ref{tab:citation_acc}. This demonstrates that \textit{learning universal representations scales well to graphs of larger size.}

\paragraph{Adaptability:} Finally, we study the adaptability of U-SSL models to new datasets. We examine if the representations learnt from a set of graphs can be used to solve the downstream task for a new graph. Here, we start with the model obtained with U-SSL of the $5$ smaller citation networks -- that has $5$ graph-specific modules $\{\mathbf{\Theta_i}\}$, $i\in[1,5]$. We leverage the modularity of the U-SSL model, and introduce a new graph-specific module $\mathbf{\Theta_6}$ dedicated to the new graph, \texttt{OGBN-arxiv}, keeping the URL module $\mathbf{\Phi}$ unchanged. We perform self-supervision with the new dataset and learn only $\mathbf{\Theta_6}$, in effect learning to project the node features of the new dataset to the common representation space. The adapted model achieves a classification accuracy of $0.538 \pm 0.002$, comparable to that of the U-SSL model trained with $6$ datasets ($0.54 \pm 0.002$), and still approximately $8$ points better than training a new model from scratch with SSL, demonstrating the adaptability of U-SSL models. Thus, one can train a U-SSL model with a set of benchmark datasets, and then simply learn a graph-specific module for a new dataset to achieve comparable performance. This prevents repetitive self-supervision for new graphs as they are made available, and is a remarkable feature of the framework that \textit{enables re-use of the learnt representations, thereby reducing the computational cost of building universal models.}

\subsection{U-SSL accommodates multiple pretext tasks}
In this section, we demonstrate the ability of U-SSL to accommodate multiple pretext tasks while delivering the above benefits. We use the pair-wise node distance (\textbf{PairDis}) as an additional pretext task for self-supervision. Here, the network is trained to predict the pair-wise distances between a pair of nodes. We consider the five citation datasets as earlier, and construct the U-SSL model with $5$ graph-specific encoders, $1$ URL module and $2$ task-specific heads for pre-training. We use the loss function described in Equation \ref{eq:loss_generalised} to tune the parameters $\mathbf{\Theta_i}$, $\mathbf{\Phi}$ and $\mathbf{\Gamma_j}$, $\forall$ $i\in[1,5]$ and $\forall$ $j\in[1,2]$.
\begin{equation}
    \mathcal{L}_{USSL}=\sum_{i=1}^N\sum_{j=1}^MW_j\mathcal{L}_{SSL, i, j}\left(\mathbf{X_i}, \mathbf{A_i}; \mathbf{\Theta_i}, \mathbf{\Phi}, \mathbf{\Gamma_j}\right). \label{eq:loss_generalised}
\end{equation}

The node classification accuracy of the models are shown in Table \ref{tab:two_tasks}. We can see that pre-training with two tasks results in $6\%$ improvement in performance for \texttt{CoraFull} and $2\%$ improvement for \texttt{Citeseer}. It is noteworthy that while performing self-supervised learning with multiple tasks, weighing the loss for each task is typically performed to achieve an improvement in performance. However, we have not performed a search for the optimal weights ($W_j$ in Equation \ref{eq:loss_generalised}), and have assigned equal weights to both the tasks, i.e., $W_1=W_2=1$. Even with this configuration, we obtain performance improvements for two datasets. These results support the general effectiveness of our framework in improving the performance of features learnt through self-supervised learning. Future studies will be aimed at improving optimising the weights of different tasks to achieve consistent improvement in performance.
\begin{table}[]
    \centering\caption{Node classification accuracy of supervised baseline, SSL and U-SSL models for citation datasets, pretrained on one and two pretext tasks. Entries in boldface represent best performance.}
    \label{tab:two_tasks}
    \begin{tabular}{ccc}
        \toprule
         Dataset & U-SSL (1 task) & U-SSL (2 tasks) \\
         \hline
         \texttt{CoraFull} & $0.60$ & $\mathbf{0.66}$ \\
         \texttt{Cora-ML} & $\mathbf{0.84}$ & $0.81$ \\
         \texttt{DBLP} & $\mathbf{0.83}$ & $0.81$ \\
         \texttt{Citeseer} & $0.86$ & $\mathbf{0.88}$ \\
         \texttt{PubMed} & $\mathbf{0.87}$ & $0.86$ \\
         \bottomrule
    \end{tabular}
\end{table}

\subsection{U-SSL generalises to multiple families}
To demonstrate the generalisability of U-SSL for multiple graph families, we compare the performance of SSL, U-SSL and supervised baselines for the co-purchase family with \texttt{computers}, \texttt{photo} datasets, and social networks with \texttt{}, . The downstream node classification performance for the co-purchase graphs are shown in Table \ref{tab:copurchase_acc}.
\begin{table}[]
    \centering\caption{Node classification accuracy of supervised baseline, SSL and U-SSL models for co-purchase datasets. Entries in boldface represent best performance out of SSL and U-SSL.}
    \label{tab:copurchase_acc}
    \begin{tabular}{cccc}
        \toprule
         Dataset & Baseline & SSL & U-SSL \\
         \hline
         \texttt{computers} & $0.90 \pm 0.007$ & $0.83 \pm 0.001$ & $\mathbf{0.86} \pm 0.001$ \\
         \texttt{photo} & $0.94 \pm 0.004$ & $0.91 \pm 0.001$ & $\mathbf{0.92} \pm 0.001$ \\
         \bottomrule
    \end{tabular}
\end{table}
We obtain $3\%$ improvement for \texttt{computers}, and $1\%$ improvement for \texttt{photo}. This shows that U-SSL can learn generalisable features for diverse families of graphs, while exhibiting less training time and using only $54\%$ of parameters compared to SSL. The performance for social network datasets are shown in Table \ref{tab:social_acc}. We achieve a performance improvement of $3\%$ each for \texttt{ego-Facebook} and \texttt{Flickr}. We also observe that the U-SSL model outperforms the supervised baseline for \texttt{Flickr} (also better than SSL) and \texttt{Twitch} (comparable with SSL). These results further demonstrate that, relying on the underlying repeating patterns in graphs of a family, U-SSL generalises to multiple families of graphs.

\begin{table}[]
    \centering\caption{Node classification accuracy of supervised baseline, SSL and U-SSL models for social network datasets. Entries in boldface represent best performance out of SSL and U-SSL. Underlined entries represent U-SSL outperforming supervised baseline.}
    \label{tab:social_acc}
    \begin{tabular}{cccc}
        \toprule
         Dataset & Baseline & SSL & U-SSL \\
         \hline
         \texttt{ego-Facebook} & $0.93 \pm 0.005$ & $0.86 \pm 0.003$ & $\mathbf{0.89} \pm 0.003$ \\
         \texttt{Facebook} & $0.93 \pm 0.004$ & $0.89 \pm 0.002$ & $0.89 \pm 0.002$ \\
         \texttt{Flickr} & $0.48 \pm 0.05$ & $0.48 \pm 0.003$ & $\underline{\mathbf{0.51}} \pm 0.002$ \\
         \texttt{Twitch} & $0.63 \pm 0.008$ & $0.69 \pm 0.002$ & $\underline{0.69} \pm 0.001$ \\
         \bottomrule
    \end{tabular}
\end{table}

\section{Outlook}
This article reports the first attempt at developing a framework to learn from multiple graphs and shows that there is computational (training efficiency and model size) and performance benefits to be gained. This paper opens up numerous potential directions for further improvement.

\paragraph{Limitations and future work} We used a naive configuration of multiple pretext task learning that assigns equal weight to both tasks. It has been shown that optimising the parameters can provide a boost in performance \citet{jin2022automated}, which can be developed further to achieve more significant performance gains. We consider only node classification as the downstream task, and a multitude of tasks, e.g., link prediction, graph classification can be considered in the future. 
Finally, the current framework unifies learning across graphs of a family, but still needs a distinct head for each pretext task. Future work can be directed to address this and unify learning across graphs and tasks, paving the way for more powerful foundation graph models.

\paragraph{Broader impact} Current research in representation learning is advancing the field towards artificial general intelligence, with foundation models and multi-modal training being major developments in this direction. These models learn representations from different types of data sources, e.g., images, videos and text, that are generalizable across multiple datasets, and at times, across multiple tasks. This work is aligned along these lines, and proposes a framework to build graph foundation models, and learn universal features from multiple graphs.

\section{Conclusion}
This work studies the problem of learning universal features across graphs of a family through self-supervision. We present a novel universal SSL framework that constructs foundation model with multiple graph-specific encoders and one universal representation learning module. Specifically, we employ graph-specific encoders to homogenize disparate features from multiple graphs, and the universal module to learn generic representations from the homogenized features. We construct one U-SSL model with a state-of-the-art graph transformer, and with extensive experiments, show that the proposed framework provides an efficacious, efficient, scalable and adaptable approach to learn universal representations from graphs.

\bibliographystyle{plainnat}
\bibliography{refs}


\appendix

\section{Implementation details}\label{app:implementation}
All experiments are performed on an NVIDIA DGX-A100 Workstation with four A100 GPUs, each with 40 GB memory. Software is implemented using PyTorch Geometric software library. The implementation of pair-wise attribute similarity is adapted from the implementation of \cite{jin2022automated}. The official implementation of NAGphormer \cite{chen2023nagphormer} is used to construct the URL module of all models. The Adam optimizer is used to learn the parameters of all models. The base learning rate is set to $1e^{-3}$ for pre-training and supervised learning, and $1e^{-2}$ for fine-tuning of SSL and U-SSL models. A learning rate scheduler that reduces the learning rate when the loss does not decrease for $50$ epochs is employed. Self-supervision is performed for $2500$ epochs, and fine-tuning is performed for $1000$ epochs for SSL and U-SSL models. Supervised baseline models are trained for $500$ epochs.

\section{Ablation study with citation networks} \label{app:ablation}
In this section, we present the ablation analsyes for the main results (citation networks).

\subsection{Depth vs width of URL module}
The ablation study of U-SSL model with respect to the dimension of transformer embedding is reported in Table \ref{tab:ablation_dim}. As expected, the performance of the model consistently decreases with smaller embedding dimension for all datasets. The results of ablation with respect to the transformer depth are reported in Table \ref{tab:ablation_depth}. Contrary to Table \ref{tab:ablation_dim}, we observe that the performance of the model does not necessarily increase with greater depth of the URL module. In fact, for all datasets except \texttt{CoraFull}, increasing the depth of the URL module from $4$ to $6$ results in poorer performing model. This suggests that the expressive power, and hence performance of the models is more reliant on having high-dimensional embeddings than a deep URL module.
\begin{table}[]
    \centering
    \caption{Ablation results with respect to transformer embedding size. Entries in boldface represent best performance.}
    \label{tab:ablation_dim}
    \begin{tabular}{cccc}
        \toprule
         Dataset & \multicolumn{3}{c}{Transformer embedding size} \\
         \cline{2-4}
         & $256$ & $128$ & $64$ \\
         \hline
         CoraFull & $\mathbf{0.60} \pm 0.003$ & $0.56 \pm 0.002$ & $0.52 \pm 0.002$ \\
         Cora-ML & $\mathbf{0.84} \pm 0.001$ & $0.77 \pm 0.003$ & $0.78 \pm 0.002$ \\
         DBLP & $\mathbf{0.83} \pm 0.001$ & $0.81 \pm 0.002$ & $0.80 \pm 0.001$ \\
         Citeseer & $\mathbf{0.86} \pm 0.002$ & $0.82 \pm 0.002$ & $0.77 \pm 0.001$ \\
         PubMed & $\mathbf{0.87} \pm 0.001$ & $0.85 \pm 0.001$ & $0.84 \pm 0.007$ \\
         \bottomrule
    \end{tabular}
\end{table}

\begin{table}[]
    \centering
    \caption{Ablation results with respect to transformer depth. Entries in boldface represent best performance.}
    \label{tab:ablation_depth}
    \begin{tabular}{cccc}
        \toprule
         Dataset & \multicolumn{3}{c}{Transformer depth} \\
         \cline{2-4}
         & $2$ & $4$ & $6$ \\
         \hline
         CoraFull & $0.61 \pm 0.002$ & $0.60 \pm 0.003$ & $\mathbf{0.77} \pm 0.001$ \\
         Cora-ML & $0.83 \pm 0.003$ & $\mathbf{0.84} \pm 0.001$ & $0.82 \pm 0.003$ \\
         DBLP & $\mathbf{0.83} \pm 0.002$ & $\mathbf{0.83} \pm 0.001$ & $0.80 \pm 0.003$ \\
         Citeseer & $0.85 \pm 0.003$ & $\mathbf{0.86} \pm 0.002$ & $0.82 \pm 0.002$ \\
         PubMed & $0.86 \pm 0.001$ & $\mathbf{0.87} \pm 0.001$ & $0.85 \pm 0.001$ \\
         \bottomrule
    \end{tabular}
\end{table}

\subsection{Architecture of URL module}
The accuracies of GNN models and the GTX model for citation networks are shown in Table \ref{tab:ablation_architecture}. The quantities in parentheses represent the improvement in performance of U-SSL models with respect to SSL models. The GTX model has comparable performance for \texttt{CoraFull}, $1\%$ lower performance for \texttt{Cora-ML} and better performance for \texttt{DBLP} ($2\%$), \texttt{Citeseer} ($2\%$) and \texttt{PubMed} ($1\%$). We observe that the GCN model does not provide any improvement in accuracy for four out of five datasets, and provides an improvement of $3\%$ for \texttt{PubMed}. On the other hand, GraphSAGE provides improvements of $1\%$ each for \texttt{CoraML} and \texttt{Citeseer} datasets, while exhibiting $2\%$ fall in performance for \texttt{DBLP}. The NAGphormer-based U-SSL model provides consistent improvement in performance for all datasets, and also outperforms the GNN-based models for majority of the datasets. Thus, the transformer-based U-SSL model provides a better modeling approach to learn universal representations across graphs.
\begin{table}[]
    \centering
    \caption{Ablation results with respect to architecture of universal representation learning module. Entries in parentheses represent the improvement compared to SSL models. Entries in boldface represent best performance.}
    \label{tab:ablation_architecture}
    \scalebox{0.85}{
    \begin{tabular}{ccccc}
        \toprule
         Dataset & \multicolumn{4}{c}{URL architecture} \\
         \cline{2-5}
         & GTX & GCN & GraphSAGE & GAT\\
         \hline
         CoraFull & $0.47 \pm 0.003 (0.00)$ & $0.60 \pm 0.004 (-0.006)$ & $0.55 \pm 0.004 (-0.001)$& $0.45 \pm 0.003 (-0.161)$ \\
         Cora-ML & $0.78 \pm 0.001 (-0.01)$ & $\mathbf{0.86} \pm 0.002 (-0.005)$ & $0.82 \pm 0.004 (0.01)$ & $0.74 \pm 0.002 (-0.107)$ \\
         DBLP & $\mathbf{0.82} \pm 0.001 (0.02)$ & $0.80 \pm 0.002 (-0.008)$ & $0.81 \pm 0.002 (-0.02)$ & $0.79 \pm 0.002 (-0.034)$ \\
         Citeseer & $\mathbf{0.82} \pm 0.002 (0.02)$ & $0.85 \pm 0.002 (0.0001)$ & $0.84 \pm 0.002 (0.01)$ & $0.81 \pm 0.003 (-0.039)$ \\
         PubMed & $\mathbf{0.83} \pm 0.001 (0.01)$ & $0.86 \pm 0.002 (0.03)$ & $0.83 \pm 0.002 (-0.005)$ & $0.84 \pm 0.002 (-0.006)$ \\
         \bottomrule
    \end{tabular}
    }
\end{table}

\section{Ablation with graphs from multiple families}\label{app:multiple-families}
In the previous results, we consider graphs belonging to one family (citation networks or co-purchase networks) and show that U-SSL learns better features than SSL. We also investigate if including graphs from more than one family also results in better performance. To achieve this, we perform combined training with all the $5$ citation networks and $2$ co-purchase networks, and summarise the results in Table \ref{tab:two_families}. We observe that out of the $7$ datasets, the performance of U-SSL is better (in comparison to SSL) for $1$ dataset, worse for $4$ datasets, and unchanged for $2$ dataset. Based on these results, we cannot claim that U-SSL can always learn better representations when trained across multiple families of graphs. This result corroborates the reasoning behind our hypothesis, i.e., graphs of the same family exhibit commonalities, and thus a combined learning framework can leverage the underlying common patterns to improve the performance.
\begin{table}[]
    \centering\caption{Node classification accuracy of supervised baseline, SSL and U-SSL models for citation and co-purchase datasets. Entries in boldface represent best performance out of SSL and U-SSL. Underlined entries represent U-SSL models that match supervised baseline performance.}
    \label{tab:two_families}
    \begin{tabular}{ccc}
        \toprule
         Dataset & U-SSL (2 families) & U-SSL (1 family) \\
         \hline
         \texttt{CoraFull} & $0.60 \pm 0.003$ & $\mathbf{0.60} \pm 0.003$ \\
         \texttt{Cora-ML} & $\mathbf{0.85} \pm 0.002$ & $0.84 \pm 0.001$ \\
         \texttt{DBLP} & $0.81 \pm 0.001$ & $\mathbf{\underline{0.83}} \pm 0.001$ \\
         \texttt{Citeseer} & $0.85 \pm 0.004$ & $\mathbf{0.86} \pm 0.002$ \\
         \texttt{PubMed} & $0.86 \pm 0.002$ & $\mathbf{\underline{0.87}} \pm 0.001$ \\
         \texttt{computers} & $0.85 \pm 0.001$ & $\mathbf{0.86} \pm 0.001$ \\
         \texttt{photo} & $\mathbf{\underline{0.92}} \pm 0.001$ & $0.92 \pm 0.001$ \\
         \bottomrule
    \end{tabular}
\end{table}


\newpage
\section*{NeurIPS Paper Checklist}

\begin{enumerate}

\item {\bf Claims}
    \item[] Question: Do the main claims made in the abstract and introduction accurately reflect the paper's contributions and scope?
    \item[] Answer: \answerYes{} 
    \item[] Justification: Section 5

\item {\bf Limitations}
    \item[] Question: Does the paper discuss the limitations of the work performed by the authors?
    \item[] Answer: \answerYes{} 
    \item[] Justification: Section 6

\item {\bf Theory Assumptions and Proofs}
    \item[] Question: For each theoretical result, does the paper provide the full set of assumptions and a complete (and correct) proof?
    \item[] Answer: \answerNA{} 
    \item[] Justification: The paper is experimental and does not contain any theoretical components.

    \item {\bf Experimental Result Reproducibility}
    \item[] Question: Does the paper fully disclose all the information needed to reproduce the main experimental results of the paper to the extent that it affects the main claims and/or conclusions of the paper (regardless of whether the code and data are provided or not)?
    \item[] Answer: \answerYes{} 
    \item[] Justification: Experiment details are provided in Section 4 and implementation details are provided in Appendix A.

\item {\bf Open access to data and code}
    \item[] Question: Does the paper provide open access to the data and code, with sufficient instructions to faithfully reproduce the main experimental results, as described in supplemental material?
    \item[] Answer: \answerYes{} 
    \item[] Justification: The data used in the expriments are standard datasets available on PyTorch Geometric or on SNAP websites. Code has been anonmymised and attached as supplementary material on OpenReview. Code will be made openly available via GitHub after review process.

\item {\bf Experimental Setting/Details}
    \item[] Question: Does the paper specify all the training and test details (e.g., data splits, hyperparameters, how they were chosen, type of optimizer, etc.) necessary to understand the results?
    \item[] Answer: \answerYes{} 
    \item[] Justification: Implementation details are provided in Appendix A.

\item {\bf Experiment Statistical Significance}
    \item[] Question: Does the paper report error bars suitably and correctly defined or other appropriate information about the statistical significance of the experiments?
    \item[] Answer: \answerYes{} 
    \item[] Justification: All experimental results presented in the paper are based on $10$ runs of training. The results are reported in \texttt{mean $\pm$ std.dev.} format.

\item {\bf Experiments Compute Resources}
    \item[] Question: For each experiment, does the paper provide sufficient information on the computer resources (type of compute workers, memory, time of execution) needed to reproduce the experiments?
    \item[] Answer: \answerYes{} 
    \item[] Justification: Implementation details with the workstation and GPU details are provided in Appendix A.
    
\item {\bf Code Of Ethics}
    \item[] Question: Does the research conducted in the paper conform, in every respect, with the NeurIPS Code of Ethics \url{https://neurips.cc/public/EthicsGuidelines}?
    \item[] Answer: \answerYes{} 
    \item[] Justification: 

\item {\bf Broader Impacts}
    \item[] Question: Does the paper discuss both potential positive societal impacts and negative societal impacts of the work performed?
    \item[] Answer: \answerYes{} 
    \item[] Justification: Broader impacts are discussed in Section 6.
    
\item {\bf Safeguards}
    \item[] Question: Does the paper describe safeguards that have been put in place for responsible release of data or models that have a high risk for misuse (e.g., pretrained language models, image generators, or scraped datasets)?
    \item[] Answer: \answerNA{} 
    \item[] Justification: The paper does not deal with LLMs or any form of generative AI.

\item {\bf Licenses for existing assets}
    \item[] Question: Are the creators or original owners of assets (e.g., code, data, models), used in the paper, properly credited and are the license and terms of use explicitly mentioned and properly respected?
    \item[] Answer: \answerYes{} 
    \item[] Justification: All codes that have been re-used and/or adapted to generate the results for the ppaer have been cited.

\item {\bf New Assets}
    \item[] Question: Are new assets introduced in the paper well documented and is the documentation provided alongside the assets?
    \item[] Answer: \answerNA{} 
    \item[] Justification: No new assets are created in this paper.

\item {\bf Crowdsourcing and Research with Human Subjects}
    \item[] Question: For crowdsourcing experiments and research with human subjects, does the paper include the full text of instructions given to participants and screenshots, if applicable, as well as details about compensation (if any)? 
    \item[] Answer: \answerNA{} 
    \item[] Justification: Crowdsourcing and research with human subjects have not been conducted in this work.

\item {\bf Institutional Review Board (IRB) Approvals or Equivalent for Research with Human Subjects}
    \item[] Question: Does the paper describe potential risks incurred by study participants, whether such risks were disclosed to the subjects, and whether Institutional Review Board (IRB) approvals (or an equivalent approval/review based on the requirements of your country or institution) were obtained?
    \item[] Answer: \answerNA{} 
    \item[] Justification: No human participants were involved in obtaining the results for the paper.

\end{enumerate}

\end{document}